\documentclass{article}

\PassOptionsToPackage{numbers}{natbib}
\usepackage[preprint]{neurips_2020}

\usepackage[utf8]{inputenc} 
\usepackage[T1]{fontenc}    
\usepackage{hyperref}       
\usepackage{url}            
\usepackage{booktabs}       
\usepackage{amsfonts}       
\usepackage{nicefrac}       
\usepackage{microtype}      
\usepackage{graphicx}
\usepackage{amsmath}
\usepackage{amsthm}
\usepackage{amssymb}
\usepackage{multirow}
\usepackage{chngpage}

\title{MC-BERT: Efficient Language Pre-Training via a Meta Controller}

\author{%
  Zhenhui Xu\thanks{Equal contribution. Works done while interning at Microsoft Research Asia.} \\
  Peking University \\
  \texttt{zhenhui.xu@pku.edu.cn} \\
  \And
  Linyuan Gong\footnotemark[1] \\
  Peking University \\
  \texttt{gonglinyuan@hotmail.com} \\
  \And
  Guolin Ke \\
  Microsoft Research \\
  \texttt{Guolin.Ke@microsoft.com} \\
  \And
  Di He \\
  Peking University \\
  \texttt{dihe@microsoft.com} \\
  \And
  Shuxin Zheng \\
  Microsoft Research \\
  \texttt{Shuxin.Zheng@microsoft.com} \\
  \And
  Liwei Wang \\
  Peking University \\
  \texttt{wanglw@cis.pku.edu.cn} \\
  \And
  Jiang Bian \\
  Microsoft Research \\
  \texttt{jiang.bian@microsoft.com} \\
  \And
  Tie-Yan Liu \\
  Microsoft Research \\
  \texttt{Tie-Yan.Liu@microsoft.com} \\
}

\begin{document}

\maketitle

\begin{abstract}
Pre-trained contextual representations (e.g., BERT) have become the foundation to achieve state-of-the-art results on many NLP tasks. However, large-scale pre-training is computationally expensive. ELECTRA, an early attempt to accelerate pre-training, trains a discriminative model that predicts whether each input token was replaced by a generator. Our studies reveal that ELECTRA's success is mainly due to its reduced complexity of the pre-training task: the binary classification (replaced token detection) is more efficient to learn than the generation task (masked language modeling). However, such a simplified task is less semantically informative. To achieve better efficiency and effectiveness, we propose a novel meta-learning framework, MC-BERT. The pre-training task is a multi-choice cloze test with a reject option, where a meta controller network provides training input and candidates. Results over GLUE natural language understanding benchmark demonstrate that our proposed method is both efficient and effective: it outperforms baselines on GLUE semantic tasks given the same computational budget.
\end{abstract}

\section{Introduction}

In natural language processing, pre-trained contextual representations are widely used to help downstream tasks without  sufficient labeled data. Previous works~\citep{radford2019language, yang2019xlnet, devlin2018bert,liu2019roberta} train contextual language representations on self-supervised generation tasks. For example, BERT~\citep{devlin2018bert} randomly masks\footnote{In BERT, among all tokens to be predicted, 80\% of tokens are replaced by the \texttt{[MASK]} token, 10\% of tokens are replaced by a random token, and 10\% of tokens are unchanged.} a small subset of the unlabeled input sequence and trains a generator to recover the original input. Such tasks require only unlabeled free texts, and \citet{Raffel_Shazeer_Roberts_Lee_Narang_Matena_Zhou_Li_Liu_2019} shows that a large dataset is crucial to a pre-trained model's performance. Pre-training over such large-scale data consumes huge computational resources, which raises a critical concern in terms of high energy cost~\citep{strubell2019energy}.

ELECTRA \citep{clark2019electra} is a successful attempt to boost the efficiency of pre-training. The learning framework of ELECTRA consists of a discriminator and a generator. Given a sentence, it corrupts the sentence by replacing some words with plausible alternatives sampled from the generator. Then, the discriminator is trained to predict whether a word in the corrupted sentence was replaced by the generator. Finally, the learned discriminator will be used in downstream tasks. Unlike previous generation tasks where the model makes predictions only on a small number (e.g., 15\% in BERT) of masked positions, the discriminative task proposed in ELECTRA is defined over all input tokens. According to \citet{clark2019electra}, this approach has better sample efficiency and, consequently, accelerated training.

In Section~\ref{sec:electra}, we provide empirical studies on ELECTRA, showing that ELECTRA has a vital advantage of reducing the complexity of the pre-training task: the \emph{replaced token detection} task of ELECTRA is a simple binary classification. It is easier to learn than generation tasks (i.e., predicting one word from the entire vocabulary), such as \emph{masked language modeling (MLM)} used by BERT. We trained two variants of ELECTRA. We first replace the simple discriminative task by a more complex task, and this modification significantly slows down the convergence. Then, we train discriminator only on a sampled subset of positions, and the convergence is not impacted significantly. These empirical studies show that for efficient training, reduced task complexity is much more important than sample efficiency.
Still, the replaced token detection task of ELECTRA is less informative than generation tasks. The semantic information required to detect replaced tokens is not as much as recovering the original input. Detailed analysis on GLUE natural language understanding benchmark shows that ELECTRA's advantage over BERT is less significant on semantic-related tasks than on syntax-related tasks.

\begin{figure}[t]
    \centering
    \includegraphics[width=\textwidth]{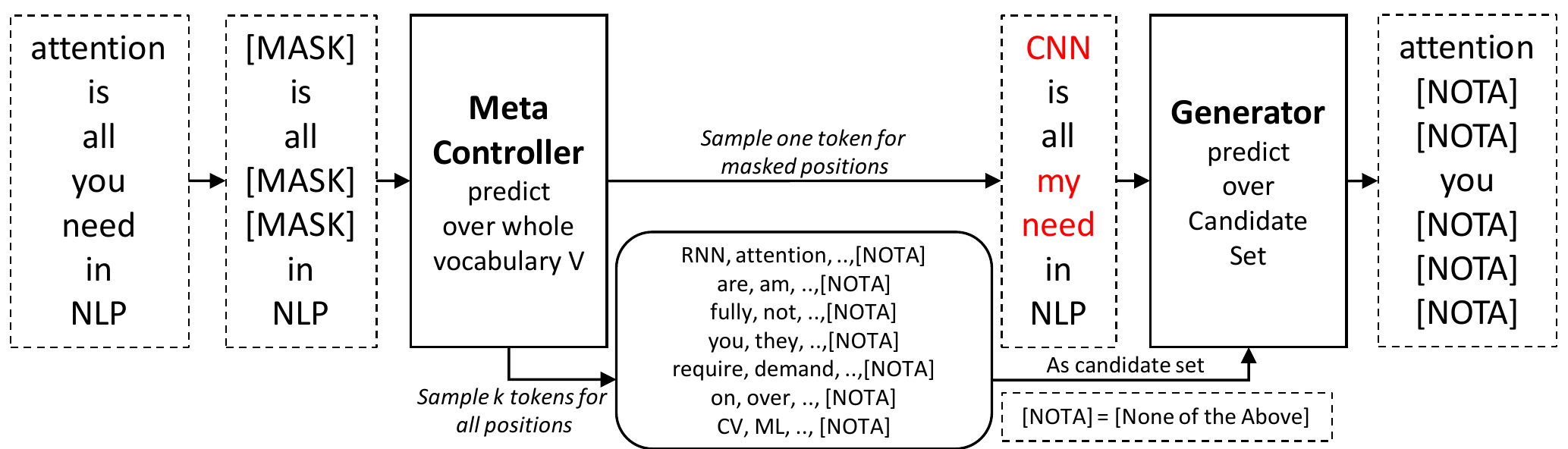}
    \caption{
    The learning framework of MC-BERT.
    Given a sentence, the meta controller first corrupts the sentence by replacing a small subset of tokens with sampled plausible alternatives. It then creates $k$ token candidates for each position. The generator uses the corrupted sentence as input and learns to correct each word by predicting over the $k$ candidates \citep{doi:10.1177/107769905303000401}.
    }
    \label{fig:overview}
\end{figure}

In Section~\ref{sec:method}, we propose \textbf{MC-BERT}, a novel language pre-training method using a \textbf{M}eta \textbf{C}ontroller to manage the training of a generator, as shown in Figure~\ref{fig:overview}. This pre-training task is comparable to multiple-choice cloze tests. Unlike BERT, the MC-BERT generator only needs to make a $k$-way classification, which reduces the task complexity. Unlike ELECTRA, MC-BERT still trains a generator, learning more semantic information.


In Section~\ref{sec:experiments}, to compare with other models, we conduct experiments and evaluate them over GLUE natural language understanding benchmark \citep{DBLP:journals/corr/abs-1804-07461}. Results show that MC-BERT is more efficient and achieves better accuracy than other baselines on most of the semantic understanding tasks.

\section{Background}

Current state-of-the-art natural understanding systems learn pre-trained contextual representations by encoding the word's surrounding context. The encoders are trained by self-supervised tasks using large-scale unlabeled corpora. For instance, \citet{peters2018deep, radford2018improving} train language models using LSTMs \citep{hochreiter1997long} or Transformer decoders \citep{vaswani2017attention}, and use the hidden states in the networks as the contextual representation. \citet{devlin2018bert,liu2019roberta} use the \emph{masked language modeling} task and achieve state-of-the-art performance on natural language understanding tasks. Alternatively, XLNet \citep{yang2019xlnet} and UniLM models \citep{dong2019unified} design permuted and bidirectional language modeling tasks.

The exploding demand of computations, together with the resulting massive energy cost~\citep{strubell2019energy}, has become an obstacle to the application of pre-training. Unfortunately, to the best of our knowledge, there is a limited number of works aiming at improving the training efficiency of such models.
\citet{you2019large} attempts to accelerate BERT pre-training, but it has to pay back with massive computational resources.
\citet{gong2019efficient} observes that parameters of BERT in different layers have structural similarity and reduce training time using implicit parameter sharing. A notable improvement is ELECTRA \citep{clark2019electra}, 
the starting point of our work. We will discurss ELECTRA in detail in Section~\ref{sec:electra}.

\section{A Deep Dive into ELECTRA} \label{sec:electra}
ELECTRA consists of a generator network $G$ and a discriminator network $D$, both of which use Transformer encoders as their backbone. Formally,
we use $V$ to denote the vocabulary of tokens; we use $x = (x_1, ..., x_n)$ to denote a sentence of $n$ tokens, where $x_i\in V$, $i=1,2,\cdots,n$; $x^{M}=\text{Mask}(x, p)$ denotes a masked sentence of $x$ in which the MASK operator randomly replaces the token at each position by a mask symbol \texttt{[MASK]} with an equal probability $p$. 

Let $x^{M}$ be the input, at each masked position, the generator $G$ learns to predict the correct token from the vocabulary: for any masked position $i$ in $x^M$, let $P(v | x^{M},i;G,V)$ be the probability that $G$ predicts $v\in V$ as the missing token, satisfying $\sum_{v \in V}P(v | x^{M},i;G,V)=1$. We use $P(\cdot | x^{M},i;G,V)$ to denote this probability distribution over $V$. The generator $G$ is trained to minimize the MLM loss as
\begin{equation} \label{eq:mlm}
\begin{aligned}
L_{\text{MLM}}(x; G) = \mathbb{E}(\sum_{i: x^{M}_i=\texttt{[MASK]}} - \log P (x_i | x^{M},i;G,V)
),\end{aligned}
\end{equation}
where the expectation is taken over the random draw of masked positions. Other details of the generator $G$ can be referred to in~\citet{devlin2018bert,clark2019electra}. 

In ELECTRA, the generator $G$ predicts the missing tokens and fill the corresponding masked positions, but the predictions may differ from the original sentence. We denote the sentence generated by $G$ as $x^{R}=\text{Replace}(x^M, G)$, in which each token $x^{R}_i, (i=1,\cdots,n)$ is defined as
\begin{equation}\label{eq:ele_replace}
x^{R}_i=\text{Replace}(x^M, G)=\left\{
             \begin{array}{lr}
             x^M_i, \text{if } x^M_i \neq \texttt{[MASK]}. &  \\
             v \sim P(\cdot | x^{M},i;G,V), \text{if } x^M_i = \texttt{[MASK]}. & \\
             \end{array}
\right.
\end{equation}

The discriminator $D$ learns to classify whether each token in $x^{R}$ is the same as the original one. To achieve this, $D$ uses Transformer encoder to get the contextual representations $(h_1,\cdots,h_n)$, where $h_i\in R^d$ is a $d$-dimension contextual embedding for position $i$. Then, $D$ introduces a binary classifier with parameters $w \in R^d$, to decide the probability of whether $x^R_i$ is the same as the original one, i.e.,
\begin{equation} \label{eq:sigmoid}
\begin{aligned}
P(x^R_i \text{ is original}; D) = \text{sigmoid}(w^T h_i),
\end{aligned}
\end{equation}

The learning objective of $D$ is to minimize the classification error, formally
\begin{equation}\label{eq:disc}\begin{aligned}
&L_{\text{DISC}}(x, x^R;D) \\
=& \mathbb{E}(-\sum_{i=1}^{n} [\mathbf{1}(x_i^R = x_i) \log P(x^R_i \text{ is original}; D) \\
+& \mathbf{1}(x_i^R \neq x_i) \log (1 - P(x^R_i \text{ is original}; D))])
\end{aligned}\end{equation}
The generator $G$ and discriminator $D$ are jointly optimized according to Eq.~\ref{eq:mlm} and~\ref{eq:disc}. After training, the discriminator $D$ will be used in downstream tasks. 

\subsection{The Real Advantage of ELECTRA over BERT}
\citet{clark2019electra} claims that ELECTRA yields higher training efficiency than BERT due to \textit{higher sample efficiency}. While the MLM loss (Eq.~\ref{eq:mlm}) of BERT is calculated over a sampled masked subset of positions (e.g., 15\%), the loss of the discriminator in ELECTRA (Eq.~\ref{eq:disc}) is calculated over all input positions.
Therefore, the learning signals enclosed in more positions could be used to optimize the model parameters, resulting in more efficient training.

However, there is another critical difference between BERT and ELECTRA: BERT learns to predict the correct word from the entire vocabulary $V$, whose size is tens of thousands. On the contrary, ELECTRA's discriminator learns from a much simpler pre-training task, i.e., predicting whether each word is replaced or not. The \textit{reduced task complexity} may also lead to training acceleration.

Given the above two crucial differences between ELECTRA and BERT, we conduct controlled experiments to examine which of them is more critical for efficient training.

\paragraph{Experimental setup} We conduct experiments to analyze the effects of higher sample efficiency or reduced task complexity on training efficiency. We use the same dataset, model architectures, and other hyperparameters as ELECTRA-Base \citep{clark2019electra}. The pre-trained models are evaluated on GLUE benchmark (\textbf{G}eneral \textbf{L}anguage \textbf{U}nderstanding \textbf{E}valuation) \citep{DBLP:journals/corr/abs-1804-07461}. We leave detailed experiment setups in Section~\ref{sec:exprimentdesign}.

To study the effects of sample efficiency, we design a modified version of ELECTRA, called \textit{ELECTRA-sample}. Unlike ELECTRA that calculates the loss of $D$ over all input positions, ELECTRA-sample only calculates the loss over 50\% of input positions (all masked positions plus a sampled subset of non-masked positions). ELECTRA-sample has lower sample efficiency than the original ELECTRA, but it keeps the same task complexity. If sample efficiency is essential to training efficiency, we can expect ELECTRA-sample's worse performance compared to ELECTRA.

To study whether a more complex pre-training task will reduce ELECTRA's training efficiency, we design a modified version of ELECTRA, called \textit{ELECTRA-complex}. Instead of training the model to check whether each word in a corrupted sentence is replaced, ELECTRA-complex learns to predict the correct word from the entire vocabulary at each position. If the task simplification is essential for ELECTRA's success, we can expect much slower training by ELECTRA-complex.

\paragraph{Results}
As we study training efficiency, we focus on each model's performance of the first several epochs. For all experiments, we dump four checkpoints at 20k, 50k, 100k, 200k steps, corresponding to 2\%, 5\%, 10\%, 20\% of all pre-training steps. All checkpoints are then fine-tuned on three downstream tasks, CoLA, RTE, and STS-B.

\begin{figure}[htb]
\centering
\includegraphics[width=\textwidth]{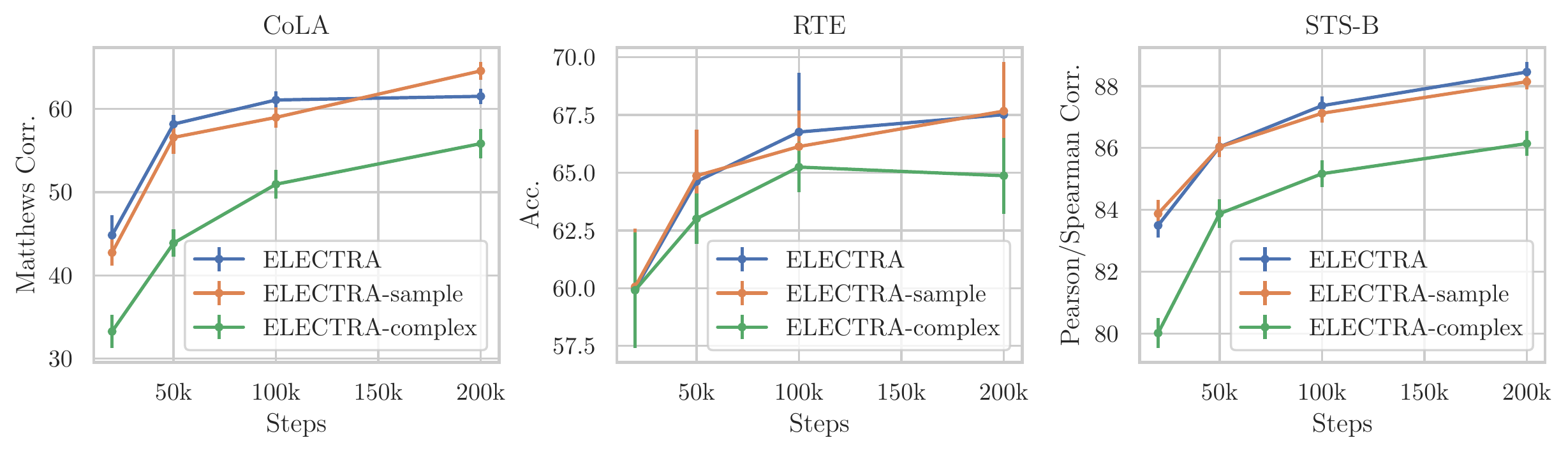}
\caption{Performance of modified ELECTRA models on downstream tasks.}
\label{fig:ele_modify}
\end{figure}

From Figure~\ref{fig:ele_modify}, we can see that ELECTRA-sample's performance is only slightly worse than ELECTRA in most of the checkpoints, although its sample efficiency is halved. This fact indicates that sample efficiency has little impact on the performance of the model.

However, from Figure~\ref{fig:ele_modify}, we can see that ELECTRA-complex's performance is consistently worse than ELECTRA by a large margin in almost every checkpoint. This fact indicates that reducing task complexity is important to improving pre-training efficiency.

\paragraph{Drawbacks of the discriminative task} It is worth noting that the discriminative task is not as informative as the generation task. Formally, we denote random variable $X$ as a sentence with any underlying distribution and $X^{R}$ as the corrupted sentence. We define a binary vector $Z$ where $Z_i=I[X_i=X^R_i]$. $Z$, therefore, is the target of the discriminative task. We have the conditional entropy $H(Z|X,X^R)=0$ since $Z$ is a deterministic function of $X$ and $X^R$. Then, it is straightforward to see that $H(X|X^R)=H(X,Z|X^R)>H(Z|X^R)$.

Empirical results in Table~3 of \citet{clark2019electra} and in Table~\ref{tab:main_res} of this paper also show that ELECTRA's advantage over BERT mainly lies in syntactic tasks (CoLA) instead of semantic tasks, which require the model to capture richer semantic information. These facts inspire us to design more informative pre-training tasks beyond ELECTRA.

\section{Pre-training with a Meta Controller\label{sec:method}}

In this section, we introduce a novel pre-training method, \textit{MC-BERT}. We still pre-train a generator (instead of a discriminator) to learn more semantic information, but we use a \textit{meta controller} to improve its training efficiency. We continue to use all notations defined in Section~\ref{sec:electra} in this section.

\subsection{MC-BERT}
Our method trains two Transformer encoders, a generator $G_{\text{model}}$ and a meta controller $G_{\text{ctrl}}$. The generator $G_{\text{model}}$ is served as the primary model and will be further used in the downstream tasks, while the meta controller guides the training of the generator.

The meta controller $G_{\text{ctrl}}$ is trained using the MLM loss defined in Eq.~\ref{eq:mlm}. Given an input sentence $x$, the meta controller guides the training of the generator $G_{\text{model}}$ in two ways:

\begin{itemize}
    \item Similar to ELECTRA, $G_{\text{ctrl}}$ generates an corrupted sentence $x^{R}=\text{Replace}(x^M, G_{\text{ctrl}})$ as is shown in Eq.~\ref{eq:ele_replace}
    \item $G_{\text{ctrl}}$ creates a set of token candidates for each position $\bar{V}_i$, and each $|\bar{V}_i|=k$, where $k$ is a small integer.
\end{itemize}

The generator $G_{\text{model}}$ uses $x^R$ as input and learns to correct the sentence using the given candidates $\bar{V}_i$ for each position $i$. In the following, we denote $\bar{V}=(\bar{V}_1,\bar{V}_2,...,\bar{V}_n)$, the tuple of all candidate sets.

\paragraph{Label Leaking and Reject Options} It is non-trivial to construct a meaningful $\bar{V}_i$ for training $G_{\text{model}}$. First, $\bar{V}_i$ should contain useful negative candidates, which can provide $G_{\text{model}}$ with informative signals for learning. Moreover, the learning process may suffer from \textit{label leaking}. Concretely, if the ground truth token appears in the candidate set $\bar{V}_i$ of every non-replaced position, the generator $G_{\text{model}}$ can easily make correct predictions by choosing the input token, since the ground truth is always the same as the input token for a non-replaced position. Because most positions are non-replaced, this problem leads to ineffective training of $G_{\text{model}}$. However, we cannot fix this problem by removing the ground truth token from $\bar{V}_i$, since it will result in an invalid classification task, where no candidate is correct.

To address this problem, we use a novel way to construct $\bar{V}$ motivated by the history of voting \citep{feddersen1999abstention, ambrus2017case}. We introduce a special category, ``None of the above'' (\texttt{[NOTA]}), as a reject option. Given a corrupted sentence $x^R$, for position $i$, if $x^R_i = x_i$ (when position $i$ is not masked, or the prediction of $G_{\text{ctrl}}$ is correct), we sample $k-1$ negative tokens without replacement according to $G_{\text{ctrl}}$, using them together with \texttt{[NOTA]} as $\bar{V}_i$. In this case, we hope $G_{\text{model}}$ can select \texttt{[NOTA]} from $\bar{V}_i$, indicating the input token is correct. If $x^R_i \ne x_i$ (when position $i$ is masked and the prediction of $G_{\text{ctrl}}$ is wrong), we sample $k-2$ negative tokens according to $G_{\text{ctrl}}$, using them together with $\{\texttt{[NOTA]} ,x_i\}$ as $\bar{V}_i$. In this case, we hope $G_{\text{model}}$ can choose $x_i$ from $\bar{V}_i$. Formally, we construct $\bar{V}_i$ as is described below.
\begin{equation} \label{eq:v_construct}
\bar{V}_i=\left\{
             \begin{array}{lr}
             \{v_i^1,...,v_i^{k-1}\}\cup \{\texttt{[NOTA]}\}, \text{if } x^R_i=x_i.   &  \\
            \{v_i^1,...,v_i^{k-2}\}\cup \{x_i,\texttt{[NOTA]}\}, \text{if } x^R_i \neq x_i. & 
             \end{array}
\right.,
\end{equation}
All negatives $v_i^j\sim P(\cdot | x^{M},i;G_{\text{ctrl}},V)$  $(j=1,2,\cdots)$ are drawn without replacement. We use $P(\cdot | x^R, i; G_{\text{model}}, \bar{V}_i)$ to denote the output distribution of $G_{\text{model}}$ over $\bar{V}_i$. Given contextual representations $h_i$ produced by $G_{\text{model}}$, and the token embedding matrix $\hat{\text{Emb}}_i$ (including \texttt{[NOTA]}) in $\hat{V}_i$,
\begin{equation} \label{eq:g_model_output}
\begin{aligned}
P(\cdot | x^R, i; G_{\text{model}}, \bar{V}_i) = \text{Softmax}(\hat{\text{Emb}}_i^T h_i),
\end{aligned}
\end{equation}

The loss function of $G_{\text{model}}$ is defined as the negative log likelihood for a $k$-class classification problem.

\begin{equation} \label{eq:l_gmodel}
\begin{aligned}
&L_{\text{model}}(x, x^R;G_{\text{model}};\bar{V}) \\
=&\mathbb{E}(-\sum_{i=1}^{n} [\mathbf{1}(x_i^R = x_i) \log P(\texttt{[NOTA]}|x^R, i; G_{\text{model}}, \bar{V}_i) \\
+&\mathbf{1}(x_i^R \neq x_i) \log P(x_i |x^R, i; G_{\text{model}}, \bar{V}_i)]).
\end{aligned}
\end{equation}

We optimize a combined loss of Eq.~\ref{eq:mlm} and Eq.~\ref{eq:l_gmodel}:

\begin{equation}\label{eq:combine_loss}
\min_{G_{\text{model}},G_{\text{ctrl}}} L_{\text{MLM}}(x; G_{\text{ctrl}}) + \lambda L_{\text{model}}(x, x^R;G_{\text{model}};\bar{V}).
\end{equation}

\subsection{Discussions}

\begin{table}[htbp]
\centering
\caption{Example of the task comparisons between BERT/ELECTRA and our proposed MC-BERT.} 
\begin{tabular}{llll}
\toprule
\multicolumn{4}{c}{Ground Truth: \textit{He is overweight as he eats a lot.}} \\ \midrule
\textbf{Model} & \textbf{Question} &  \textbf{Choices} & \textbf{Answer} \\ \midrule
BERT & \textit{He is \underline{\ \ \ \ \ \ \ \ } as he eats a lot.} & All tokens: \textit{abandon, able, about, \dots} & \textit{overweight} \\ 
ELECTRA & \textit{He is \underline{a} as he eats a lot.} & Right, Wrong & Wrong \\ 
MC-BERT & \textit{He is \underline{tiny} as he eats a lot.} & A. \textit{overweight}\ \ \ \ \ B. \textit{healthy}   & A \\ 
 &  & C. \textit{smart}\ \ \ \ \ D. \textit{None of the above} & \\ \bottomrule
\end{tabular}
\label{tab:task}
\end{table}

The example in Table~\ref{tab:task} illustrates the difference between MC-BERT and BERT/ELECTRA in terms of their pre-training tasks. From Table~\ref{tab:task}, we can see that BERT solves a general cloze problem: it masks some tokens and requires the learner to pick correct tokens from the entire vocabulary. The task is very complex. ELECTRA learns from detecting replaced tokens, which is a binary classification problem similar to grammar checking. This task is less complex, but the learning signal of ELECTRA is less informative.

Our MC-BERT is similar to multi-choice cloze tests that have been widely used in real practices, such as the GRE verbal test. Moreover, the input sequence and the candidates are given by the meta controller network, which gradually increases the difficulty of the generator's pre-training task. In the beginning, the meta controller is not well-trained, so it provides the generator with easy multi-choice questions. Therefore, the generator can learn from these questions efficiently. As the meta controller outputs more meaningful token alternatives and negative candidates, the generator will be forced to make predictions relying on deep semantic information from contexts. In conclusion, MC-BERT strikes a good balance between training efficiency and the richness of semantic information learned by the model.

Our method is related to curriculum learning \citep{bengio2009curriculum}. Curriculum learning suggests that some instances are easier to learn, and the model training should first focus on easy instances and then on the hard ones. Our work is different from curriculum learning in that we consider the complexity of the self-supervised tasks rather than the difficulty of instances.

Note that our methodology is quite general. As the main idea is to simplify the generation task using a meta controller, it is easy to be extended to help a broad class of self-supervised pre-training methods, such as XLNet and UniLM \citep{dong2019unified}.

\section{Experiments}\label{sec:experiments}
In this section, we evaluate our proposed MC-BERT with BERT and ELECTRA on a wide range of tasks. We implement all methods based on \emph{fairseq} \cite{gehring2017convs2s} in PyTorch \footnote{Codes have been anonymously released to  \url{https://github.com/MC-BERT/MC-BERT} for review.}. For BERT, we use the implementation of RoBERTa (an optimized version of BERT) \cite{liu2019roberta} in \emph{fairseq}. We will use RoBERTa to refer to as BERT in the following of this section. 

\subsection{Experimental Setup\label{sec:exprimentdesign}}
\begin{table}[t]
    \centering 
    \caption{Hyperparameter search spaces for fine-tuning. Other hyperparameters are kept the same as pre-training.}
\begin{tabular}{lr}
\toprule
\textbf{Batch size} & \{16, 32\} \\ 
\textbf{Maximum epoch} & 10 \\ 
\textbf{Learning rate} & \{1e-5,~...,~8e-5\}  \\ 
\textbf{Warm-up ratio} & 0.06 \\ 
\textbf{Weight decay} & 0.1 \\ \bottomrule

\end{tabular}
    \label{tab:ds_space}
\end{table}

\paragraph{Model architecture} We use the same architecture for RoBERTa, the discriminator $D$ of ELECTRA, and the generator $G_{\text{model}}$ of MC-BERT, where we set all hyperparameters to be the same as BERT-Base (110M parameters). The only difference between these three models lies in the number of output categories of the output layer. \citet{clark2019electra} recommends using a small-size generator for better efficiency. For a fair comparison, we set the architecture of our meta controller $G_{\text{ctrl}}$ to be the same as the ELECTRA generator $G$.

\paragraph{Pre-training} We use the same pre-training corpus as \cite{devlin2018bert}, which consists roughly 3400M words from English Wikipedia corpus\footnote{\url{https://dumps.wikimedia.org/enwiki}} and BookCorpus\footnote{As the dataset BookCorpus \citep{moviebook} is no longer freely distributed, we follow the suggestions from \citet{devlin2018bert} to crawl from \url{smashwords.com} and collect BookCorpus by ourselves. }. We apply \textit{byte pair encoding} (BPE) \citep{DBLP:journals/corr/SennrichHB15} with the same vocabulary size as BERT, where $|V|=32768$.

We construct the inputs of MLM models (RoBERTa, the ELECTRA discriminator, and the meta controller of MC-BERT) in the same way as \citet{devlin2018bert}. For MC-BERT, we set the number of token candidates $k=20$ and set the factor of the generator's loss function $\lambda=20$, unless otherwise specified.

We use the same sequence lengths, batch sizes, and training steps as \citet{devlin2018bert} for all models. In total, we train each model for 1 million steps. We use the same optimizer configuration as \citet{liu2019roberta} and the same learning rate scheduling scheme as \citet{devlin2018bert}. We train all models on 8 NVIDIA Tesla V100 GPUs.


\paragraph{Fine-tuning} 
We use the GLUE (\textbf{G}eneral \textbf{L}anguage \textbf{U}nderstanding \textbf{E}valuation) benchmark\citep{DBLP:journals/corr/abs-1804-07461} as the downstream tasks to evaluate the performance of the pre-trained models. GLUE consists of nine tasks. CoLA is a syntactic task where the model checks the linguistic acceptability of each sentence. Other tasks, such as SST-2 (sentiment analysis), STS-B (semantic text similarity), and MNLI (natural language inference), are semantic tasks. The detailed description of each task is shown in the supplementary materials.

We run each configuration with ten different random seeds and take the average of these ten scores as the performance of this configuration. We report the best score over all configurations.

\begin{table}[ht]
    \centering
    \caption{The results on the GLUE benchmark. The percentage numbers of the FLOPs denote the progress of pre-training.}
    \begin{tabular}{llrrrrrr}
    \toprule
        \multirow{2}*{\textbf{Task}} & \multirow{2}*{\textbf{Model}} & \multicolumn{6}{c}{\textbf{Pre-train FLOPs}}  \\ 
        ~ & ~ & 4\% & 8\% & 16\% & 32\% & 64\% & 100\% \\ \midrule
        \multirow{2}*{Syntactic} & RoBERTa & 27.21 & 42.23 & 47.00 & 50.69 & 57.40 & 57.41 \\
        \multirow{2}*{(CoLA)} & ELECTRA & \textbf{44.83} & \textbf{58.15} & \textbf{61.05} & \textbf{61.49} & \textbf{65.72} & \textbf{64.34} \\
        ~ & MC-BERT & 39.20 & 53.27 & 57.96 & 59.20 & 62.05 & 62.10 \\ \midrule
        \multirow{2}*{Semantic} & RoBERTa & 76.40 & 80.06 & 81.83 & 82.85 & 84.41 & 84.65 \\
        \multirow{2}*{(8 tasks)} & ELECTRA & 79.57 & 82.76 & 84.22 & 85.23 & 86.15 & 86.52 \\
         ~ & MC-BERT & \textbf{79.78} & \textbf{83.23} & \textbf{84.28} & \textbf{85.46} & \textbf{86.63} & \textbf{86.82} \\
    \bottomrule
    \end{tabular}
    \label{tab:main_res}
\end{table}

\subsection{Experiment Results}

To compare efficiency fairly, we define a list of computational costs (in terms of FLOPs). For each experiment, we dump the checkpoint trained with respective computational cost and then fine-tune it in downstream tasks. All corresponding results are shown in Table 
~\ref{tab:main_res}.

\paragraph{Syntactic tasks} Table~\ref{tab:main_res} shows that our proposed MC-BERT is significantly better than RoBERTa under different computational constraints, which indicates that MC-BERT is much more efficient than RoBERTa. On the other hand, for this particular task, CoLA, ELECTRA outperforms both RoBERTa and MC-BERT, because the pre-training task of ELECTRA is more aligned with CoLA. As we discussed in Section~\ref{sec:electra} and Section~\ref{sec:method}, the \textit{replaced token detection} pre-training task of ELECTRA mainly provides the model with syntactic information, making it do particularly well in syntactic acceptability tasks. Therefore, we would like to focus on the comparison of different models on the other eight tasks, which are semantic tasks that require deeper semantic understanding.

\begin{figure}[t]
    \centering
    \includegraphics[width=\textwidth]{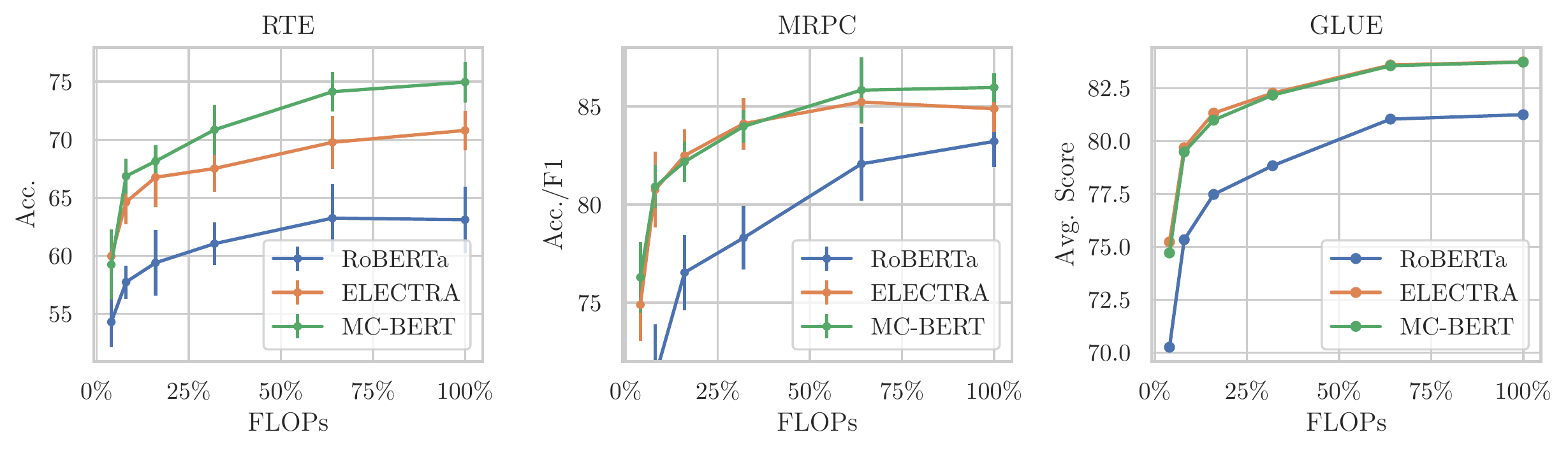}
    \caption{Left: Model performances on RTE; Middle: Model performances on MRPC; Right: GLUE scores.}
    \label{fig:glue}
\end{figure}

\paragraph{Semantic tasks} We report the average performance of each checkpoint on the eight tasks. As shown in Table~\ref{tab:main_res}, MC-BERT consistently outperforms RoBERTa and ELECTRA in almost all checkpoints, which indicates that MC-BERT is more efficient than RoBERTa and ELECTRA in learning semantic information from texts. In Figure~\ref{fig:glue} (Left and Middle), we show the learning curves of two semantic tasks, RTE and MRPC. For both tasks, MC-BERT achieves higher performance than ELECTRA and RoBERTa under the same computational budgets. For tasks that require deeper semantic understanding, our proposed MC-BERT has more significant advantages in terms of efficiency and effectiveness than the baselines do.

\paragraph{Discussion} The above experimental results show that MC-BERT outperforms BERT on all the tasks, indicating the effectiveness of using a meta controller to help the generator's training. They also suggest that the generator-discriminator framework in ELECTRA is not the only way to achieve better efficiency. 

We plot the final GLUE scores of all model checkpoints in Figure~\ref{fig:glue} (Right). Our MC-BERT is competitive to ELECTRA in terms of the average performance of the nine tasks. However, MC-BERT does better in eight semantic tasks but performs worse on the one syntactic task.

\subsection{Effect of hyper-parameters}

\begin{figure}
    \centering
    \includegraphics[width=170pt]{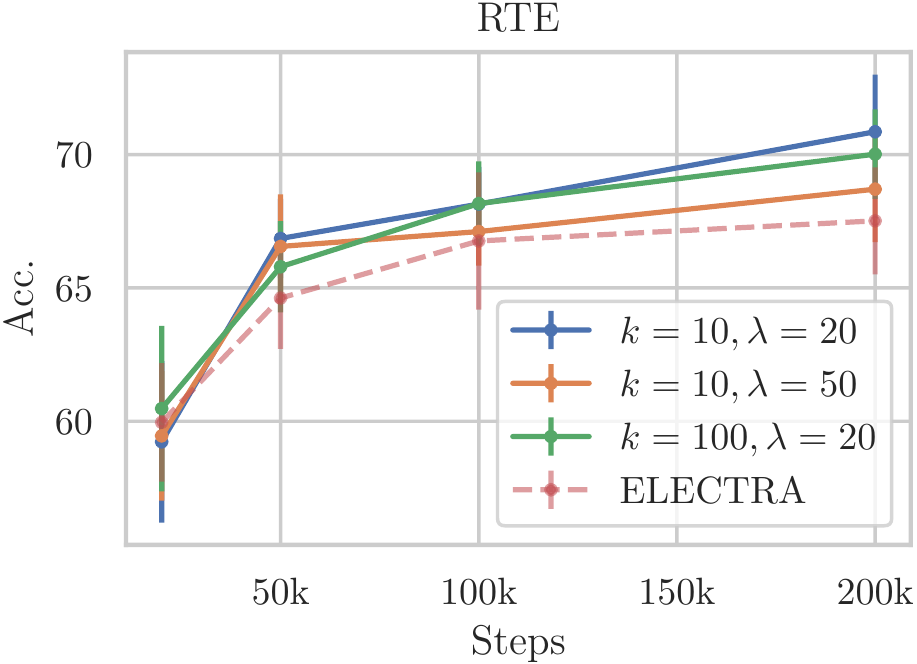}
    \caption{Effect of hyper-parameters in MC-BERT.}
    \label{fig:fig_sensitivity}
\end{figure}

We examine the effect of hyper-parameters used in MC-BERT. We follow the experimental settings described above and assess the pre-trained models' performance on the RTE task. The experimental results are shown in Figure~\ref{fig:fig_sensitivity}.

\paragraph{Effect of varying $k$} If $k$ is very large, e.g., $k\approx |V|$, MC-BERT will be comparable to ELECTRA-complex (see Section~\ref{sec:electra}), so degraded performance is expected. In Figure~\ref{fig:fig_sensitivity}, we compare the performance given reasonable smaller values of $k$, specifically $k=10$ and $k=100$. The models trained with $k=10$ perform slightly better than those trained with $k=100$, but the difference is insignificant.

\paragraph{Effect of varying $\lambda$} Since $\lambda$ serves as a trade-off between learning the meta controller and the generator, a larger $\lambda$ indicates a greater focus on optimizing the generator rather than optimizing the meta controller. To check the effects of varying $\lambda$, Figure~\ref{fig:fig_sensitivity} compares the models trained with $\lambda=20$ and $\lambda=50$. We can see that the models trained with $\lambda=20$ is consistently better than the model trained with $\lambda=50$, which implies that too large $\lambda$ may hurt the performance due to the less optimized meta controller.

\section{Conclusion and Future Work}
In this work, we propose MC-BERT, which uses a meta controller to manage the complexity of the pre-training task. The pre-training task is a multi-choice cloze test with a reject option, ``None of the above''. Extensive experiments demonstrate MC-BERT is more efficient than BERT, and it learns deeper semantic information than ELECTRA does. It outperforms several baselines on semantic understanding tasks given the same computational budget. We will continue exploring more roles of the meta controller, e.g., how to mask positions and select batched sentences smartly.
\newpage

\small

\bibliographystyle{plainnat}
\bibliography{references}

\newpage
\appendix

We released the source code of MC-BERT at: \url{https://github.com/MC-BERT/MC-BERT}. Our code is based on PyTorch, Fairseq\footnote{\url{https://github.com/pytorch/fairseq}}, and RoBERTa \cite{liu2019roberta}.

We run all our experiments on NVIDIA Tesla V100. For pre-training, we use mixed-precision floating-point arithmetic (FP16+FP32) to accelerate training; for fine-tuning, we train with single-precision floating-point arithmetic (FP32). Besides, we measure the FLOPs in the same way as \citet{clark2019electra} do.

\section{Model Details}

The architecture settings for RoBERTa, ELECTRA and MC-BERT are listed in Table~\ref{tab:arch}. For the encoder served for downstream tasks, we use the same architecture for three models. As for the generator of ELECTRA, we use the same model size as \citet{clark2019electra}. We also set the size of the meta controller of MC-BERT to be the same as the size of the ELECTRA generator.

\begin{table}[hbtp]
    \centering
    \caption{Model specifications. The ``Encoder" denotes the transformer served for downstream tasks in each model, with the same base architecture for all the models. The ``$G_{\text{ctrl}}$ / $G$" denotes the controller of MC-BERT and the generator of ELECTRA, respectively, and they are also set to be the same. }
    \begin{tabular}{lrr}
       \toprule 
       \textbf{Hyperparameter} & \textbf{Encoder} & \textbf{$G_{\text{ctrl}}$ / $G$} \\ 
       \midrule
       Number of layers & 12 & 12 \\
       Hidden size & 768 & 256 \\
       FFN inner hidden size & 3072 & 1024 \\
       Attention heads & 12 &  4 \\
       Attention head size & 64 & 64 \\
       Embedding size & 768 & 768 \\
         \bottomrule
    \end{tabular}
    \label{tab:arch}
\end{table}

\section{Pre-Training Details}

\subsection{Dataset}
We use the same dataset as the one in BERT \cite{devlin2018bert}, which includes BooksCorpus and Wikipedia. After concatenating these two datasets, we obtain a corpus with roughly 3400M words in total. Following the practices of \citet{devlin2018bert}, we first segment documents into sentences with Spacy\footnote{\url{https://spacy.io}}; then, we normalize, lower-case, and tokenize texts using Moses decoder \citep{Koehn2007MosesOS}; next, we apply \textit{byte pair encoding} (BPE) \citep{DBLP:journals/corr/SennrichHB15}. We randomly split documents into one training set and one validation set, where the training-validation ratio for pre-training is 199:1. The vocabulary consists of 32,768 tokens. Following \citet{liu2019roberta}, we pack each input with full sentences sampled contiguously from the corpus, such that the total length is at most 512 tokens.

\subsection{Hyperparameters}

The pre-training hyperparameters are set mostly the same as the ones in BERT \cite{devlin2018bert}. However, as are suggested by recent works \cite{yang2019xlnet} \cite{liu2019roberta} \cite{lan2019albert}, we remove the next sentence prediction (NSP) pre-training task. The details are listed in Table~\ref{tab:pre_detail}.

\begin{table}[hbtp]
    \centering
    \caption{Pre-training hyperparameter settings.}
    \begin{tabular}{lr}
       \toprule
       \textbf{Hyperparameter} & \textbf{Pre-training Value} \\
       \midrule
       Learning rate & 1e-4 \\
       Learning rate decay & Linear \\
       Decay steps & 1,000,000 \\
       Warmup steps & 10,000 \\
       Adam $\epsilon$ & 1e-6 \\
       Adam $(\beta1, \beta2)$ & (0.9, 0.98) \\
       Batch size & 256 \\
       Dropout & 0.1 \\
       Attention dropout & 0.1 \\
       Weight decay & 0.01 \\
       ELECTRA $\lambda$ & 50 \\
       MC-BERT $\lambda$ & 20 \\
       MC-BERT $k$ & 10 \\
       \bottomrule
    \end{tabular}
    \label{tab:pre_detail}
\end{table}

\section{Down-Stream Details}
\subsection{GLUE Tasks}
We use the GLUE (\textbf{G}eneral \textbf{L}anguage \textbf{U}nderstanding \textbf{E}valuation) dataset \citep{DBLP:journals/corr/abs-1804-07461} as the downstream tasks to evaluate the performance of the pre-trained models. Particularly, there are nine tasks within the GLUE dataset that have been widely used for evaluation, including CoLA, RTE, MRPC, STS-B, SST-2, QNLI, QQP, and MNLI-m/mm. The specifications of these tasks are listed in Table~\ref{tab:glue}. Especially, we follow BERT \cite{devlin2018bert} and ELECTRA \cite{clark2019electra} to skip WNLI in our experiments, because few submissions on the leaderboard\footnote{\url{https://gluebenchmark.com/leaderboard}} do better than predicting the majority class for this task.

Notably, we strictly adopt official metrics to evaluate the performance on GLUE tasks. However, the scores reported in ELECTRA \cite{clark2019electra} are not. Their evaluation metrics are Spearman correlation for STS-B (instead of the average of Spearman correlation and Pearson correlation), Matthews correlation for CoLA, and accuracy for all the other GLUE tasks (instead of the average of F1-score and accuracy for MRPC and QQP).

\begin{table}[hbtp]
    \centering
    \caption{Specification of GLUE tasks.}
    \begin{tabular}{lrrrrr}
       \toprule
       \textbf{Corpus} & \textbf{Size} & \textbf{Task} & \textbf{\#Class} & \textbf{Metric(s)} & \textbf{Domain} \\
       \midrule
       \multicolumn{6}{c}{Syntactic Tasks} \\
       \midrule
       CoLA & 8.5k & Acceptibility & 2 & Matthews correlation & Misc. \\
       \midrule
       \multicolumn{6}{c}{Semantic Tasks} \\
       \midrule
       RTE & 2.5k & Inference & 2 & Accuracy & Misc. \\
       MRPC & 3.7k & Paraphrase & 2 & Accuracy/F1 & News \\
       STS-B & 5.7k & Similarity & - & Pearson/Spearman corr. & Misc. \\
       SST-2 & 67k & Sentiment & 2 & Accuracy & Movie reviews \\
       QNLI & 108k & QA/Inference & 2 & Accuracy & Wikipedia \\
       QQP & 364k & Similarity & 2 & Accuracy/F1 & Social QA questions \\
       MNLI-m/mm & 393k & Inference & 3 & Accuracy & Misc.\\
       \bottomrule
    \end{tabular}
    \label{tab:glue}
\end{table}

\subsection{Fine-Tuning Details}
For fine-tuning, most hyperparameters are also the same as in BERT \cite{devlin2018bert}. We design an exhaustive search for batch size, learning rate to get reasonable performance numbers. The details of the search space has been shown in the paper. Our search space is much larger than the setting in both BERT \cite{devlin2018bert} and ELECTRA \cite{clark2019electra}, with higher confidence. Except for the hyperparameters listed in space, the other parameters are all set the same as in pre-training.

\begin{table}[hbtp]
\begin{adjustwidth}{-1in}{-1in}
    \centering
    \caption{The detailed results on the GLUE benchmark (except WNLI). 
    }
\begin{tabular}{ll|r|rrrrrrr|r}
\toprule
 \multirow{2}*{\textbf{FLOPs}} & \multirow{2}*{\textbf{Model}} & \textbf{CoLA} & \textbf{SST-2} & \textbf{MRPC} & \textbf{STS-B} & \textbf{QQP} & \textbf{MNLI\footnotesize{-m/mm}} & \textbf{QNLI} & \textbf{RTE} & \textbf{Avg.} \\ & &8.5k  & 67k & 3.7k & 5.7k & 364k & 393k & 108k & 2.5k & - \\ \midrule
 \multirow{3}*{4\% (2e18)} & RoBERTa & 27.21 & 87.58 & 63.04 & 80.62 & 86.76 & 76.93/77.69 & 85.22  & 54.28 &  76.40 \\
 ~ & ELECTRA & 44.83 & 87.72 & 74.89 & 83.50 & 87.38 & 77.48/78.08 & 85.75  & 59.96 & 79.57 \\
 ~ & MC-BERT & 39.20 & 88.50 & 76.29 & 83.54 & 87.31 & 77.68/78.27 & 85.63 & 59.23 &  79.89 \\ \midrule
 \multirow{3}*{8\% (4e18)} & RoBERTa & 42.23 & 90.70 & 71.22 & 85.08 & 88.15 & 79.71/79.78 & 87.82 & 57.72 & 80.06 \\
 ~ & ELECTRA & 58.15 & 90.15 & 80.75 & 86.04 & 88.64 & 80.63/80.93 & 88.32 & 64.62 & 82.76 \\
 ~ & MC-BERT & 53.28 & 91.11 & 80.91 & 86.11 & 88.51 & 80.80/80.95 & 88.23 & 66.85 & 83.23 \\ \midrule
 \multirow{3}*{16\% (8e18)} & RoBERTa & 47.00 & 91.56 & 76.53 & 86.22 & 88.68 & 81.36/81.55 & 88.98 & 59.38 & 81.83 \\
 ~ & ELECTRA & 61.05 & 91.56 & 82.50 & 87.37 & 89.14 & 82.32/82.28 & 89.90  & 66.76 & 84.22 \\
 ~ & MC-BERT & 57.96 & 91.96 & 82.18 & 86.93 & 89.11 & 82.04/82.30 & 89.46 & 68.14 & 84.28 \\ \midrule
 \multirow{3}*{32\% (1.6e19)} & RoBERTa & 50.69 & 92.22 & 78.30 & 86.72 & 89.13 & 82.61/82.41 & 90.05 & 61.03 & 82.85 \\
 ~ & ELECTRA & 61.49 & 92.31 & 84.12 & 88.46 & 89.57 & 83.85/83.87 & 90.80 & 67.51 & 85.23 \\
 ~ & MC-BERT & 59.20 & 92.67 & 83.98 & 87.31 & 89.37 & 83.69/83.58 & 90.37 & 70.85 & 85.46 \\ \midrule
 \multirow{3}*{64\% (3.2e19)} & RoBERTa & 57.40 & 93.08 & 82.07 & 87.72 & 89.31 & 84.40/84.47 & 91.04  & 63.24 & 84.41 \\
 ~ & ELECTRA & 65.72 & 92.82 & 85.22 & 88.79 & 89.99 & 85.42/84.80 & 91.31 & 69.77 & 86.15 \\
 ~ & MC-BERT & 62.05 & 92.41 & 85.83 & 88.23 & 89.75 & 85.11/84.73 & 91.15 & 74.13 & 86.63 \\ \midrule
 \multirow{3}*{100\% (5e19)} & RoBERTa & 57.41 & 93.15 & 83.22 & 88.14 & 89.24 & 84.69/84.63 & 91.02 & 63.10 & 84.65 \\
 ~ & ELECTRA & 64.33 & 93.38 & 84.88 & 89.10 & 89.96 & 86.00/85.29 & 91.85 & 70.80 & 86.52 \\
 ~ & MC-BERT & 62.10 & 92.34 & 85.96 & 88.01 & 89.65 & 85.68/85.24 & 91.34 & 74.96 & 86.82 \\ 
  \bottomrule
\end{tabular}
    \label{tab:test_result}
\end{adjustwidth}
\end{table}

\section{Detailed Results}

Due to the space limitation, the detailed experimental results are listed here in Table~\ref{tab:test_result}. The scores on all the tasks are listed, including the scores on the nine tasks for each checkpoint. The ``Avg." is for the semantic tasks. The number below each task denotes the number of training examples. The metrics for these tasks are mentioned above. Following the standard practice for computing GLUE scores, we report the arithmetic average of all metrics for tasks with multiple metrics (MRPC, QQP, STS-B), and we average the score of MNLI-m and MNLI-mm to get the final score of MNLI.

It can be seen from the table, when the data size of the downstream task is large, e.g., in QNLI, MNLI and QQP, the performance of both RoBERTa/ELECTRA and our proposed method are similar. However, when the data size of the downstream task is small, ELECTRA and ours are significantly better than RoBERTa.

\end{document}